%% file: CameraReady.tex
\documentclass[conference]{IEEEtran}
\IEEEoverridecommandlockouts
\usepackage{cite}
\usepackage{tabularx}
\usepackage{pifont}
\usepackage{amsmath,amssymb,amsfonts}
\usepackage{algorithmic}
\usepackage{graphicx}
\usepackage{titletoc}
\usepackage{textcomp}
\usepackage{multicol,multirow}
\usepackage{bm}
\usepackage{float}
\usepackage{url}
\usepackage{xcolor}

\usepackage[table]{xcolor}
\usepackage{booktabs}
\newcommand{\pub}[1]{\color{gray}{\tiny{[{#1}]}}}
\usepackage{xcolor}
\newcolumntype{Y}{>{\centering\arraybackslash}X} 
\def\BibTeX{{\rm B\kern-.05em{\sc i\kern-.025em b}\kern-.08em
    T\kern-.1667em\lower.7ex\hbox{E}\kern-.125emX}}
\begin{document}

\title{Weakly Supervised Incremental Segmentation via Semantic Anchors and Spatial Arbitration
\thanks{Guangyu Gao is the corresponding author.
This work was supported by the National Natural Science Foundation of China (Grant 62472033, 61972036).}
}

\author{
    \IEEEauthorblockN{Zhonggai Wang, Kai Fang, and Guangyu Gao}
    \IEEEauthorblockA{
        \textit{School of Computer Science and Technology, Beijing Institute of Technology}, 
        Beijing, China \\
        q1109803861@gmail.com,
        2242452056@qq.com,
        guangyugao@bit.edu.cn
    }
}

\maketitle

\begin{abstract}
Weakly Incremental Learning for Semantic Segmentation (WILSS) suffers from the continuous introduction of noisy supervision, which progressively corrupts class-level representations, leading to severe \textit{feature drift} and semantic corruption, thereby causing newly learned classes to overwrite old ones.
To address these issues, we propose a drift-resilient WILSS approach, named \textit{SASA}, designed to stabilize semantic learning via \textit{S}emantic \textit{A}nchors and \textit{S}patial \textit{A}rbitration.
Specifically, at the representation level, we introduce semantic anchors of learnable tokens as rigid class-level references to preserve long-term semantic identity.
Complementary to this, an elastic residual adaptation facilitates controlled, instance-specific refinement, ensuring a stable yet flexible learning trajectory.
At the supervision level, we develop a Spatial Label Arbitration mechanism that performs geometry-aware decisions to directly filter unreliable signals and enforce a strict ``one object, one class'' constraint.
By synergistically stabilizing representations and improving supervision reliability, SASA effectively mitigates feature drift under weak supervision.
Extensive experiments on standard benchmarks demonstrate that our approach consistently outperforms existing state-of-the-art methods, particularly in challenging multi-step incremental settings. 
The code is available at \url{https://github.com/ZhonggaiWang/SASA}.
\end{abstract}


\section{Introduction}
\input{revised_section/intro}
\section{Proposed Method}
\input{revised_section/method}

\section{Experimental Results}
\input{revised_section/experiments}

\section{Conclusion}
\input{paper_section/conclusion}

\bibliographystyle{IEEEbib}
\bibliography{icme2026references}
\clearpage

\end{document}

%% file: revised_section/intro.tex

Incremental semantic segmentation aims to extend models to novel categories while preserving performance on previous classes~\cite{long2015fcn,cermelli2020mib}.
In realistic scenarios, dense annotations for new classes are often unavailable, motivating Weakly Supervised Incremental Semantic Segmentation (WILSS), which learns from image-level labels~\cite{cermelli2022wssil}.
Despite its practicality, WILSS is inherently challenging as noisy supervision accumulates across incremental stages, progressively distorting class-level representations and causing catastrophic forgetting.

A prevailing WILSS pipeline generates pseudo labels for novel classes using CAM~\cite{zhou2016cam} and fills uncovered regions with predictions from a frozen old model, typically combined with distillation to preserve knowledge~\cite{li2016lwf}.
While intuitive, this paradigm suffers from a fundamental limitation.
CAMs tend to activate only discriminative object parts, leaving large regions unlabeled.
To compensate, predictions from the frozen old model are utilized to pseudo-label these regions.
However, since the old model is agnostic to novel categories, it frequently misinterprets these regions as visually similar known classes rather than identifying them as background or unknown~\cite{cermelli2022wssil,teddy2024overwriting}.
Consequently, this process introduces intrinsically contradictory supervision and substantial noise within the same object instance, particularly for visually ambiguous categories.

\begin{figure}[t] 
    \centering
    \includegraphics[width=0.95\linewidth]{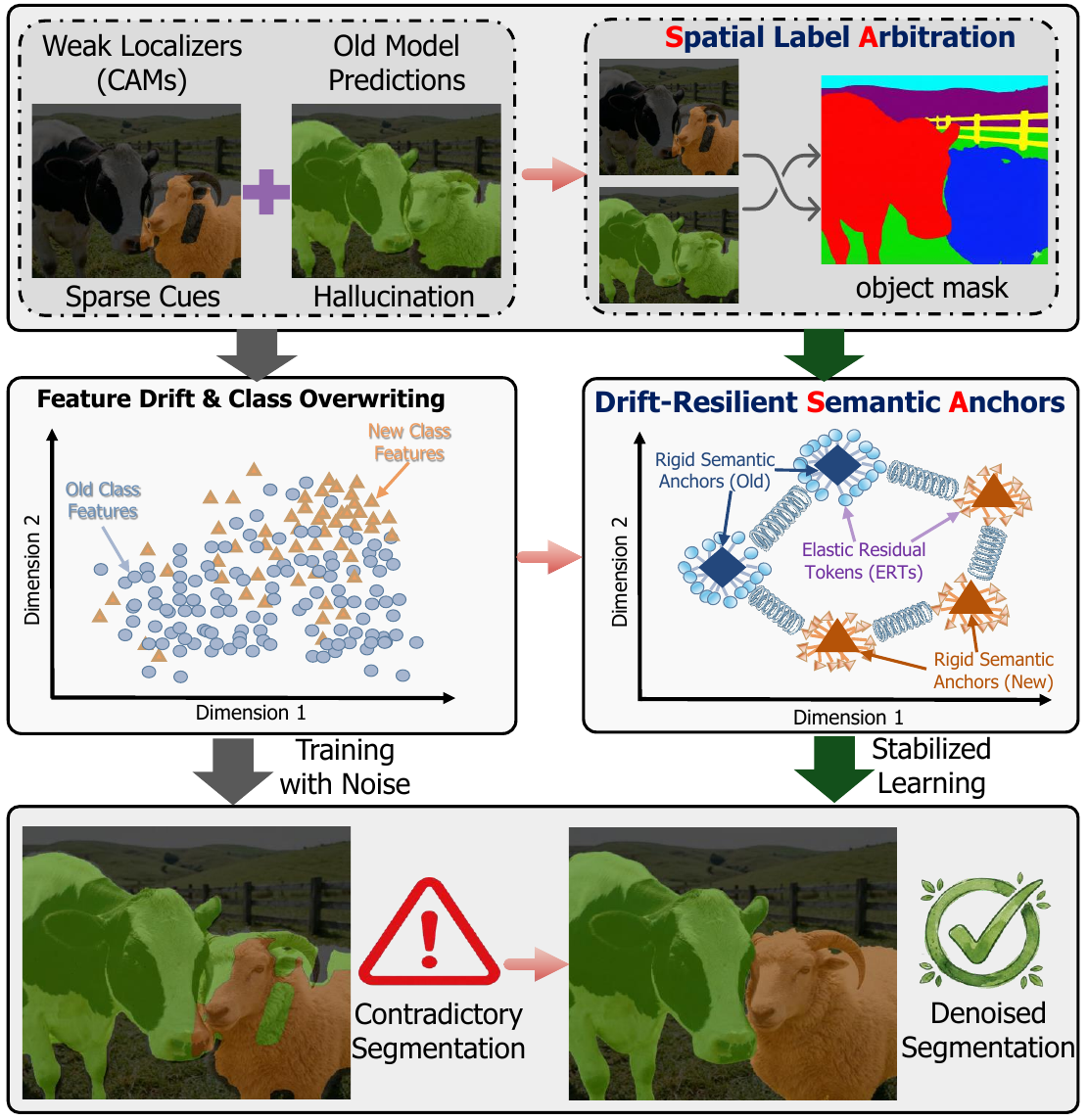}
  \caption{Illustration of Feature Drift vs. Stabilized Learning.
  (\textbf{\textcolor{darkgray}{Left}}) Merging sparse cues with hallucinations creates Contradictory Supervision, causing new classes to overwrite the old ones (\textbf{Feature Drift}). 
  (\textbf{\textcolor{green!20!black}{Right}}) We propose \textbf{Spatial Label Arbitration} to denoise labels via object masks. 
  Combined with \textbf{Rigid Anchors and Elastic Residual Tokens} (ERTs), our \textcolor{red}{SASA} maintains a stable, drift-resilient feature space throughout incremental steps.}  \label{fig:intro}
\vspace{-0.5cm}
\end{figure}

Existing efforts on WILSS can be broadly grouped into three directions.
The first focuses on enhancing weak supervision quality for novel classes, for example, by refining CAM seeds with additional priors~\cite{ahn2018affinity,ru2022attention} or leveraging foundation models to generate denser pseudo masks~\cite{yu2023foundation,liu2024l2a}.
While these alleviate sparsity in weak localization, they do not fundamentally prevent noisy signals from being repeatedly injected across incremental stages.
The second line emphasizes knowledge preservation through distillation or regularization, constraining the new model to mimic predictions or intermediate representations of the old model~\cite{li2016lwf,douillard2021plop,zhang2022dkd,zhang2022icme_incremental}.
Such strategies are effective when supervision is reliable; however, under contradictory labels, they may inadvertently propagate semantic bias and reinforce incorrect predictions.
More recently, several methods attempt to resolve conflicts at the loss or prediction level~\cite{hao2024prompt,mib20}. 
However, these approaches leave the evolution of semantic representations implicitly shaped by noisy gradients rather than being explicitly controlled.

We argue that the \textbf{pivotal challenge} of WILSS lies not merely in the imperfection of weak labels, but in the fact that persistent noise induces a biased and compounding training signal, which is only partially mitigated by existing noise-aware methods~\cite{wang2023icme_noise}.
Concretely, contradictory supervision repeatedly forces new-class features toward old-class clusters while displacing old features from their established distributions~\cite{cermelli2022wssil,mutual2024}. 
Over multiple incremental steps, this manifests as \textbf{Feature Drift}: class-level representations diverge, decision boundaries degrade, and eventually, class overwriting emerges~\cite{kemker2018survey}. 
Crucially, once the semantic reference itself becomes unstable, such drift cannot be reliably rectified by pseudo-label refinement alone.

To counteract the noise-induced drift, we adopt a simple yet principled perspective: long-term class identity should be stabilized by rigid references, while instance-level variations should be absorbed through controlled adaptations rather than being written into class semantics. 
Based on this, we propose \textit{SASA}, a drift-resilient WILSS framework based on Semantic Anchors and Spatial Arbitration.
Our approach is built upon two principled designs:
(i) \textit{Semantic Anchors with Elastic Adaptation}:
At the representation level, we maintain class-wise learnable anchors as rigid semantic references. 
Instead of allowing noisy pseudo labels to directly reshape class representations, anchors define stable coordinates in the embedding space. 
To accommodate instance-level variations without amplifying noise, an elastic residual branch performs conservative, instance-conditioned refinement around these anchors, ensuring a ``small-step'' but stable adaptation.
(ii) \textit{Spatial Label Arbitration (SLA)}: 
Since the most harmful noise in WILSS is often spatially structured (\textit{e.g.}, old/new conflicts within a single object), we incorporate a geometry-aware arbitrator. 
By leveraging these class-agnostic object masks, the arbitrator enforces a strict ``One Object, One Class'' constraint, directly filtering unreliable signals before they can corrupt the learning process.

In summary, our main contributions are threefold:
(i) We analyze WILSS from the perspective of persistent noise injection and formally characterize Feature Drift and class overwriting as compounding consequences of contradictory supervision.
(ii) We propose SASA, a framework that stabilizes incremental updates through rigid learnable anchors and elastic residual adaptation, providing a robust semantic coordinate system.
(iii) We develop a geometry-aware Spatial Label Arbitration mechanism that explicitly denoises pseudo-label supervision, further mitigating drift. Together with DSA, it addresses contradictory supervision at both the semantic and spatial levels.
Extensive experiments on standard benchmarks validate that SASA consistently outperforms existing state-of-the-art methods, especially in challenging multi-step settings.

%% file: revised_section/method.tex
\subsection{Task Definition}
Incremental learning proceeds through a sequence of training steps $\{t\}_0^T$, where the model learns series of disjoint class sets $\bm{C}^{t}$ such that $\bigcap_{t} \bm{C}^{t} = \varnothing$ and $\bigcup_{t} \bm{C}^{t} = \mathcal{\bm{C}}$. 
At each step $t$, training data $\mathcal{D}^t=\{(\mathbf{X}^{t}, \mathbf{Y}^{t})\}$ is provided.
A key characteristic of \textbf{WILSS} is \textit{asymmetric supervision}: the initial stage ($t=0$) provides dense pixel-level annotations, whereas all subsequent steps ($t>0$) provide only image-level labels $\bm{y}^{t}$ for novel classes.
Access to data from previous stages $\mathcal{D}^{0:t-1}$ is strictly prohibited.
The objective is to develop a model that effectively segments the cumulative class set $\bm{C}^{0:t}$, while simultaneously mitigating catastrophic forgetting of old categories and resisting feature drift under weak supervision.

\subsection{Drift-Resilient Semantic Anchors~(DSA)}

\subsubsection{Rigid Anchors as Class References}
To mitigate feature drift, we introduce Semantic Anchors as persistent, explicit class references in the embedding space. 
Unlike conventional prototypes computed from fluctuating features—which are highly susceptible to noise corruption—our anchors are implemented as learnable tokens $\mathbf{A}^{t} \in \mathbb{R}^{|\bm{C}^{0:t}| \times D}$. 
Each row $\mathbf{A}_k^t \in \mathbb{R}^{D}$ serves as a rigid, stable coordinate for class $k$. 
These tokens provide a persistent global framework that anchors the representations of both old and new classes. 
During training, $\mathbf{A}^{t}$ is optimized to maximize inter-class separability while maintaining long-term semantic identity, ensuring each category occupies a distinct, stable region despite the continuous influx of noisy supervision.

\begin{figure*}[t] 
    \centering    \includegraphics[width=0.95\linewidth]{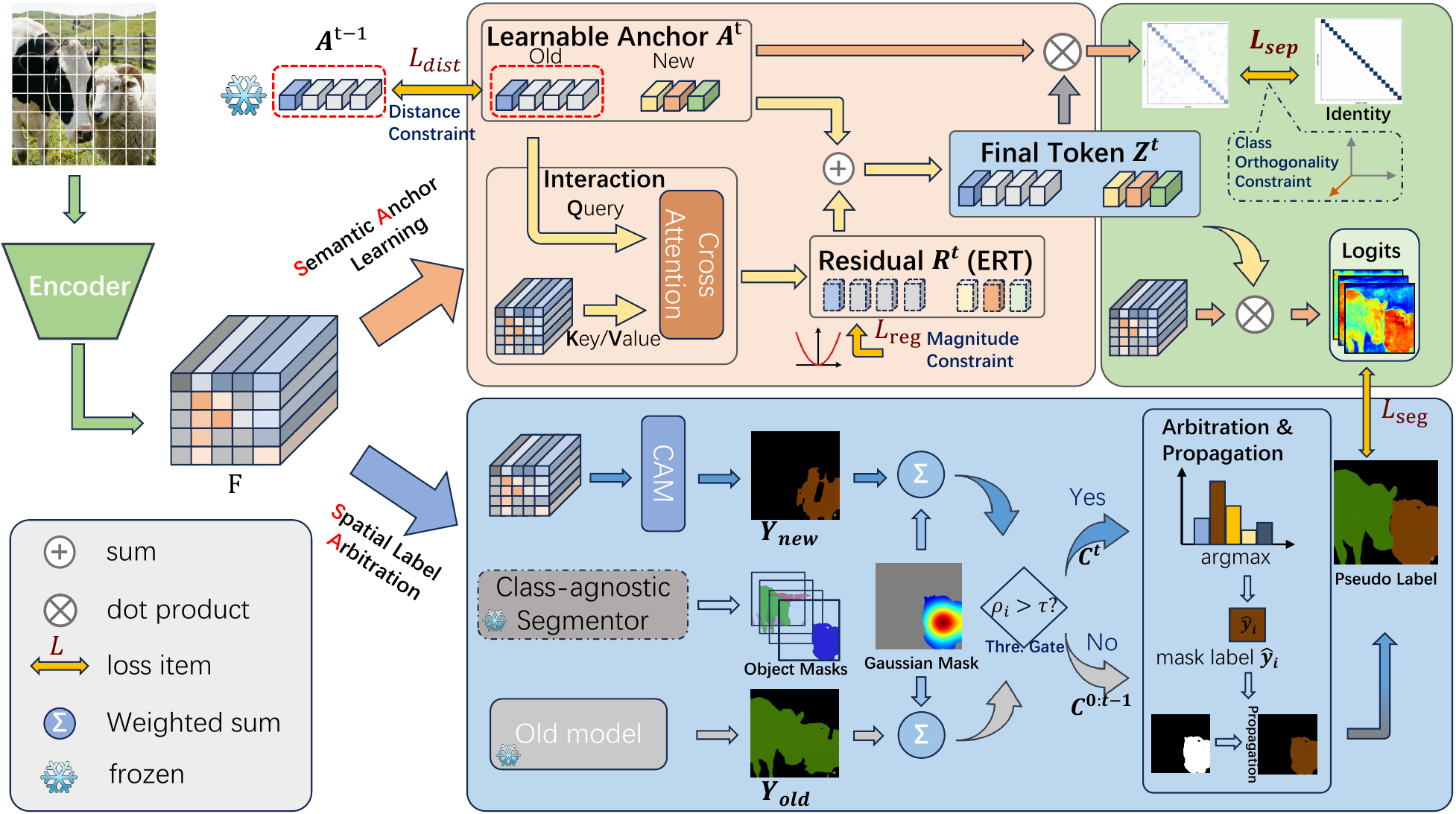}
    \caption{
    \textbf{Overview of the SASA framework.}
    (\textbf{Top}) Semantic Anchor Learning employs rigid anchors ($A^t$) and elastic residual tokens ($R^t$) under orthogonality and magnitude constraints to prevent feature drift. 
    (\textbf{Bottom}) Spatial Label Arbitration utilizes a class-agnostic segmentor to resolve supervision conflicts between $\textbf{Y}_{old}$ and $\textbf{Y}_{new}$, generating denoised pseudo-labels for spatially coherent incremental learning. 
    Segmentation logits are computed from the similarity between pixel features and class tokens in Eq. \ref{eq:similarity}, without a separate mask decoder; SLA is only used during training.
}\label{fig:framework}
\vspace{-0.3cm}
\end{figure*}

\subsubsection{Elastic Residual Tokens (ERT)}
While rigid anchors provide stability, real-world instances exhibit diverse appearance variations. 
To accommodate these without disrupting the global semantic structure, we introduce Elastic Residual Tokens (ERT).
These residual tokens, denoted as $\mathbf{R}^{t} \in \mathbb{R}^{|\bm{C}^{0:t}| \times D}$, capture instance-specific deviations from the rigid anchors. 
We dynamically generate $\mathbf{R}^{t}$ via a cross-attention mechanism where the anchors $\mathbf{A}^{t}$ act as queries and the flattened feature map $\hat{\mathbf{F}}$ serves as keys and values:
\begin{equation}
    \mathbf{R}^{t} = \text{Softmax}\left(\frac{\mathbf{A}^{t} (\mathbf{W}_K\hat{\mathbf{F}})^\top}{\sqrt{D}}\right) (\mathbf{W}_V\hat{\mathbf{F}}),
    \label{eq:ert_attention}
\end{equation}
where $\mathbf{W}_K$ and $\mathbf{W}_V$ are learnable projections.The final instance-adaptive Final Tokens (FT), denoted as $\mathbf{Z}^{t}$, are then obtained by augmenting the rigid anchors with their corresponding residuals: $\mathbf{Z}^{t} = \mathbf{A}^{t} + \mathbf{R}^{t}$. To ensure that the residuals only facilitate ``small-step'' refinements without altering core class identity, we apply an $\ell_2$ penalty: 
$\mathcal{L}_{\text{reg}} = |\mathbf{R}^t|_2^2$.

\subsubsection{Token Supervision and Regularization}
Dense prediction is performed by computing cosine similarity between pixel features $\mathbf{F}_{i,j}$ and tokens $\mathbf{Z}^t$. 
The score for class $k$ is:
\begin{equation}
    \bm{S}_{i,j,k} = \frac{1}{\tau} \left\langle \frac{\bm{Z}^{t}_k}{\|\bm{Z}^{t}_k\|_2}, \frac{\bm{F}_{i,j}}{\|\bm{F}_{i,j}\|_2} \right\rangle.
    \label{eq:similarity}
\end{equation} 
It is supervised by a CE loss $\mathcal{L}_{seg}$ using ground truth ($t=0$) or rectified pseudo-labels ($t>0$). 
This aligns $\mathbf{Z}^t$ with local embeddings while $\mathbf{A}^t$ maintains the global class center.

To prevent semantic collapse while allowing shared attributes, we propose an instance-elastic strategy instead of rigid anchor orthogonality.
We constrain FTs to align with their corresponding anchors while remaining orthogonal to others:
\begin{equation}
    \mathcal{L}_{\text{sep}} = \left\| \left( \frac{\bm{Z}^t \cdot {\bm{A}^t}^\top}{\|\bm{Z}^t\|_2 \|\bm{A}^t\|_2} \right) - \mathbf{I} \right\|_F^2.
    \label{eq:sep_loss}
\end{equation}
where $\|\cdot\|_F$ is the Frobenius norm.
Finally, to preserve old knowledge, we apply distillation on anchors of previous classes $\bm{C}^{0:t-1}$:
\begin{equation}
  \mathcal{L}_{\text{dist}}=\frac{1}{|\bm{C}^{0:t-1}|}\sum_{k\in\bm{C}^{0:t-1}}\left\|\bm{A}_k^{t} - \bm{A}_k^{t-1}\right\|_2^2.
\end{equation}
Jointly optimizing $\mathcal{L}_{\text{sep}}$ and $\mathcal{L}_{\text{dist}}$ balances plasticity for new concepts with the stability to resist catastrophic forgetting. 
This dual-constraint approach ensures the model effectively learns novel categories while preserving the integrity of previously acquired knowledge.

\subsection{Spatial Label Arbitration for Supervision Denoising~(SLA)}

In WILSS, conflicts between frozen old model predictions and noisy novel-class seeds lead to semantic fragmentation within single object instances. 
To suppress this spatially structured noise, we propose the Spatial Label Arbitration (SLA) mechanism, which enforces a ``One Object, One Class'' constraint through geometry-aware decision-making.

\subsubsection{Spatial Reliability Map}
SLA assesses pixel-level signals within class-agnostic object masks $\mathcal{M}=\{m_i\}_{i=1}^N$, generated by a general-purpose segmentor (\textit{e.g.}, SAM or Maskformer).
For each mask $m_i$ with centroid $\mathbf{c}_i$, we define a pixel-wise reliability weight $w_p$ ($p\in m_i$) to prioritize the object's ``semantic core'' over potentially noisy boundaries using a Gaussian decay function:
\begin{equation}
w_p = \exp \left( - {\| p - \mathbf{c}_i \|^2} / {2 \sigma_i^2} \right),
\end{equation}
where $\sigma_i = \alpha \sqrt{|\mathcal{A}_i|}$ is an adaptive scale based on the mask area $|\mathcal{A}_i|$. 
This formulation prioritizes the ``semantic core'' of the object near the centroid, reducing the influence of potentially noisy boundary pixels during arbitration processes.


\subsubsection{Weighted Arbitration for Label Fusion}

To resolve conflicts, we first quantify the activation density $\rho_i$ of the novel-class signal $\mathbf{Y}_{\text{new}}$ within each mask $m_i$:
\begin{equation}
    \rho_i = \frac{\sum_{p \in m_i} w_p \cdot \mathbb{I}(\mathbf{Y}_{\text{new}}(p) \in \bm{C}^{t}) }{ \sum_{p \in m_i} w_p + \epsilon },
    \label{eq:weighted_ratio}
\end{equation}
Unlike simple averaging, $\rho_i$ leverages geometric reliability to evaluate whether a mask should be treated as a novel category.
We then execute a weighted majority voting scheme to determine the final mask-level label $\hat{y}_i$:
\begin{equation}
\hat{y}_i = \arg\max_{k \in \mathcal{K}} \left( \sum_{p \in m_i} w_p \cdot \mathbb{I}(\mathbf{Y}(p) = k) \right),
\end{equation}
where the candidate set $\mathcal{K}$ is dynamically restricted on activation density: $\mathcal{K} = \bm{C}^t$ if $\rho_i > \tau$, and $\mathcal{K} = \bm{C}^{0:t-1}$ otherwise.

Upon completing the arbitration, the resolved label $\hat{y}_i$ is propagated to all pixels within the mask $m_i$, ensuring that the entire region is consistently labeled. 
These rectified pseudo-labels $\mathbf{Y}$ serve as the primary supervision for $\mathcal{L}_{seg}$, driving the joint optimization of both the instance-adaptive features and the rigid class-level anchors. 
This mechanism effectively converts fragmented, contradictory pixel signals into spatially coherent supervision for robust incremental learning.

\subsection{Overall Objective}

The total loss $\mathcal{L}$ integrates classification, segmentation, and feature-level constraints:
\begin{equation}
\mathcal{L} = \mathcal{L}_{\text{cls}} + \lambda_{\text{seg}}\mathcal{L}_{\text{seg}} + \lambda_{\text{sep}}\mathcal{L}_{\text{sep}} + \lambda_{\text{dist}} \mathcal{L}_{\text{dist}} + \lambda_{\text{reg}} \mathcal{L}_{\text{reg}},
\end{equation}
where $\mathcal{L}_{\text{cls}}$ denotes the multi-label image classification loss employed to generate CAMs.
By minimizing $\mathcal{L}$ end-to-end, SASA simultaneously optimizes instance-adaptive features and rigid class anchors while maintaining a non-overlapping, drift-resilient feature space.

%% file: revised_section/experiments.tex
\subsection{Experimental Setup}
\textbf{Datasets and Evaluation Metrics.}
We evaluate SASA on Pascal VOC 2012~\cite{pascal10}~(augmented to 10,582 images) and MS COCO~\cite{coco}~(80 categories). 
For incremental evaluation, we report mIoU for initial classes ($\bm{C}^0$), newly introduced classes ($\bm{C}^{1:T}$), and the cumulative set ($\bm{C}^{0:T}$).

\textbf{Implementation Protocols.}
We focus on the realistic overlap scenario~\cite{wilson22,teddy24}, where new classes appear alongside previously learned ones. 
On VOC, we test four settings: $15$-$5$, $10$-$10$ (2 steps), $10$-$5$ (3 steps), and $10$-$2$ (6 steps). 
We also evaluate a rigorous COCO-to-VOC setting, where 60 COCO classes are learned first, followed by incremental training on the remaining VOC classes.

\textbf{Implementation Details.}
We adopt ViT-B as the backbone.
The training utilizes a linear warm-up ($1\times10^{-6}$ to $6\times10^{-5}$) followed by a polynomial decay. 
Incremental steps use a reduced learning rate of $2\times10^{-5}$ (8k iterations on VOC; 20k on COCO-to-VOC). 
Loss weights are set as $\lambda_{seg}=0.2, \lambda_{sep}=0.2, \lambda_{dist}=0.1, \lambda_{reg}=0.05$, and the SLA confidence threshold $\tau=0.6$.

\subsection{Segmentation Results}

\begin{table*}[!t]
    \centering
    \caption{Performance comparison under different overlap settings on Pascal VOC. 
    \textit{P} and \textit{I} denote pixel-level and image-level supervision, respectively. 
    Bold indicates the best image-level results, and \underline{underline} denotes the best pixel-level performance. 
    \textit{Joint} refers to the fully-supervised upper bound. 
    \textit{(ViT)} indicates the Vision Transformer backbone, while others utilize ResNet.
    }\label{tab:voc_ss}
    \resizebox{\textwidth}{!}{
    \begin{tabular}{lr|ccc|ccc|ccc|ccc|ccc}
    \toprule
    \multirow{2}{*}{Method} & \multirow{2}{*}{Sup}  & \multicolumn{3}{c|}{\textbf{10-10 VOC}} & \multicolumn{3}{c|}{\textbf{15-5 VOC}} & \multicolumn{3}{c|}{\textbf{COCO-to-VOC}} & \multicolumn{3}{c|}{\textbf{10-2 VOC} } & \multicolumn{3}{c}{\textbf{10-5 VOC}}\\
    &  & 1-10 & 11-20 & All & 1-15 & 16-20 & All & 1-60 & 61-80 & All  & 1-10 & 11-20 & All & 1-10 & 11-20 & All\\
    \midrule
    MiB~\cite{mib20} \pub{CVPR20}& P & 70.4 &63.7 &67.2 & 75.5 &49.4 &69.0 & 34.9 &47.8 &38.7  & -& -&- & -& -&-\\
    PLOP~\cite{plop21} \pub{CVPR21}& P & 69.6 &62.2 &67.1 & 75.7 &51.7 &70.1 & 35.1 &39.4 &36.8  & -& -&- & -& -&-\\
    SDR~\cite{sdr21} \pub{CVPR21}& P & 70.5 &63.9 &67.4 & 75.4 &52.6 &69.9 & - & - & -  & -& -&- & -& -&-\\
    RECALL~\cite{recall21} \pub{ICCV21}& P & 66.0 &58.8 &63.7 & 67.7 &54.3 &65.6  & - & - & -  & -& -&- & -& -&-\\
    ALIFE~\cite{alife22} \pub{NIPS22}& P & 74.1 &\underline{69.8} &71.9 & 77.2 &52.5 &71.3 &  -& -&-&  54.8 &40.7 &48.1 &68.3& 58.8&63.8\\
    DKD~\cite{dkd22} \pub{NIPS22}& P & 75.2 &69.6 &\underline{72.5} & 78.8 &58.2 &73.9 &  41.7& 43.8&42.7& 58.7 & 45.8& 52.6&68.8& 57.6& 63.4\\
    STAR~\cite{star23} \pub{NIPS23}& P & 74.1 &68.8 &71.6 &  \underline{79.5} &\underline{58.9} & \underline{74.6}&  43.1& 44.5&44.2 & 72.3 &58.2 &65.6&73.5& 64.7&69.3\\
    BARM~\cite{background24} \pub{ECCV24}& P & \underline{76.2} &67.8 &72.2 & 78.5 &56.3 &73.2 &  \underline{44.5}&\underline{45.0} &\underline{45.3}& \underline{75.1} &\underline{59.7} &\underline{67.8} & \underline{75.7}& \underline{64.8}&\underline{70.5}\\
    \midrule
    SEAM~\cite{seam20} \pub{CVPR20}& I & 67.5 &55.4 &62.7 & 68.3 &31.8 &60.4 & 31.2 &28.2 &30.5  & -& -&- & -& -&-\\
    SS~\cite{ss20} \pub{CVPR20}& I & 69.6 &32.8 &52.5 & 72.2 &27.5 &62.1 & 35.1 &36.9 &35.5  & -& -&- & -& -&-\\
    EPS~\cite{eps21} \pub{CVPR21}& I & 69.0 &57.0 &64.3 & 69.4 &34.5 &62.1 & 34.9 &38.4 &35.8  & -& -&- & -& -&-\\
    WILSON~\cite{wilson22}  \pub{CVPR22}& I & 70.4 &57.1 &65.0 & 74.2 &41.7 &67.2 & 39.8 &41.0 &40.6  & 38.7 &22.4 &32.5 & 66.8 &46.5 &58.1\\
    FMWISS~\cite{FMWISS23} \pub{CVPR23} & I & 73.8 &62.3 &69.1 & 78.4 &\textbf{54.5} &\textbf{73.3} & 39.9 &44.7 &41.6  &  -& -&- & -& -&-\\
    Teddy~\cite{teddy24} \pub{ECCV24} & I & 71.2 &59.4 &66.5 & 77.6 &51.4 &72.0 & 40.6 &41.8 &41.5  & 50.3 &32.0 &43.1 & 68.9 &51.7 &61.7\\
    \midrule
    \rowcolor{gray!30}
    Joint (Resnet) & P & 78.4 & 76.4 & 77.4 & 79.8 & 70.2 & 77.4 & 47.8 & 46.9 & 47.7  & 78.4 & 76.4 & 77.4 & 78.4 & 76.4 & 77.4\\
    \midrule
    WILSON (ViT)~\cite{wilson22}  \pub{CVPR22}& I & 74.3 &61.2 &68.8 & 75.2 &46.4 &69.1 & 41.0 &42.3 &41.8  & 52.6 &36.1 &46.2 & 73.6 &57.5 &66.9\\
    ToCo (ViT)~\cite{toco23}  \pub{CVPR23}& I & 73.5 &58.8 &67.4 & 74.6 &44.3 &67.9 &40.3& 41.4&41.1 &49.2 &33.9 &43.4 & 72.9 &56.0 &65.7\\
    \textbf{SASA (ViT)}~\pub{Ours} & I & \textbf{76.6} & \textbf{70.1} & \textbf{74.3} & \textbf{78.7} & 52.6 & 73.0 & \textbf{47.7} & \textbf{45.1} & \textbf{47.5}  & \textbf{61.1}& \textbf{38.5} & \textbf{51.5} & \textbf{75.1}& \textbf{66.3}&\textbf{71.7}\\
    \midrule
    \rowcolor{gray!30}
    Joint(ViT) & P & 79.1 & 77.3 & 78.2 & 80.4& 71.6 & 78.2 & 50.4 & 49.5 & 50.3  & 79.1 & 77.3 & 78.2 & 79.1 & 77.3 & 78.2\\
    \bottomrule
    \end{tabular}  
    }
\end{table*}

With the synergy of Spatial Label Arbitration (SLA) and Drift-Resilient Semantic Anchors (DSA), SASA achieves substantial improvements across all settings (Table~\ref{tab:voc_ss}).

\textbf{Standard Benchmarks.}
In relatively easier settings such as 10-10 and 15-5 VOC, SASA reaches state-of-the-art performance. Specifically, under the 10-10 setting, our method improves mIoU by over 5\% for both old and new categories. In the 15-5 task, we obtain a 1.0\% overall gain. 
These improvements primarily stem from the ability of learnable anchors to stabilize the feature space against noisy supervision.

\textbf{Challenging Scenarios.}
SASA excels in complex cross-dataset and incremental settings (e.g., COCO-to-VOC and VOC 10-2) where existing approaches typically fail to maintain inter-class discriminability.
By surpassing the current SOTA by a margin of 7+ points, SASA bridges the gap toward the fully supervised counterpart, \textit{Joint-ViT}.

\textbf{Long-term Robustness.}
For the long 6-step incremental task (10-2), SASA demonstrates remarkable resilience. Compared to WILSON and Teddy, SASA improves old-class mIoU by $22.4\%$ and $10.8\%$, and by $16.1\%$ and $6.5\%$ on new classes, respectively. 
This underscores the critical role of Rigid Anchors in maintaining near-orthogonal class representations throughout a long incremental process.

\subsection{Ablation Study}

Table \ref{tab:component_ablation} presents a component-wise analysis on VOC 10-5, validating the effectiveness of DSA and SLA.
Our baseline is a plain ViT-B model using raw CAM-derived pseudo labels.

\textbf{Impact of DSA}: 
Integrating Drift-Resilient Semantic Anchors (DSA) effectively resolves feature drift. 
By maintaining rigid representations for old classes and assigning stable, non-overlapping coordinates for new ones, DSA yields a $4.7\%$ gain over the baseline. 
It ensures that the ``semantic core'' of each class remains uncorrupted by persistent noise.

\textbf{Impact of SLA}: 
The SLA module provides complementary benefits by rectifying the quality of pseudo labels at the supervision level. 
By leveraging geometry-aware object masks, SLA suppresses false positives and internal inconsistencies common in ViT architectures, contributing a $3.5\%$ improvement.

\textbf{Synergy}:
When combined, DSA and SLA lead to a total improvement of $7.6\%$ mIoU.
This confirms that stabilizing representations (DSA) and refining supervision (SLA) are complementary strategies that prevent feature drift and noise.

\begin{table}[!tbp]
\centering
\caption{Ablation study of DSA and SLA on VOC 10–5 (ViT-B/16). The default setting is highlighted in \colorbox{gray!20}{gray}.}
\scriptsize
\begin{tabular}{cc|ccc} 
    \toprule
    \textbf{DSA} & \textbf{SLA} & \textbf{Old (1–10)} & \textbf{New (11–20)} & \textbf{All} \\ 
    \midrule
     & & 71.5 & 54.2 & 64.1 \\ 
    \ding{51} & & 74.8 & 60.5 & 68.8 \\ 
     & \ding{51} & 73.0 & 59.9 & 67.6 \\
            \rowcolor{gray!20}
    \ding{51} & \ding{51} & \textbf{75.1} & \textbf{66.3} & \textbf{71.7} \\
    \bottomrule 
\end{tabular}
\label{tab:component_ablation}
\end{table}

\textbf{Sensitivity Analysis of Loss Weights $\lambda_{seg}$, $\lambda_{sep}$, and $\lambda_{dist}$.}
We evaluate $\lambda_{seg}$, $\lambda_{sep}$, and $\lambda_{dist}$ on VOC 10--5 by varying one while fixing the others at their default values ($\lambda_{seg}=0.2$, $\lambda_{sep}=0.2$, $\lambda_{dist}=0.1$). 
Table~\ref{tab:sensitivity} shows that $\lambda_{seg}=0.2$ provides the best balance, while lower values weaken supervision and higher values overfit noisy pseudo-labels. 
Performance also peaks at $\lambda_{sep}=0.2$, as weaker constraints increase inter-class confusion. 
For $\lambda_{dist}$, lower values favor plasticity but accelerate forgetting, whereas higher values overly constrain adaptation.
The default setting achieves the best overall mIoU of \textbf{71.7\%}.

\begin{table}[h]
    \centering
    \caption{Sensitivity analysis of loss weights on VOC 10--5. The default setting is highlighted in \colorbox{gray!20}{gray}.}
    \small
    \renewcommand{\arraystretch}{1.1}
    \setlength{\tabcolsep}{8pt}
    \begin{tabular}{ccc|ccc}
        \toprule
        \multirow{2}{*}{$\lambda_{seg}$} & \multirow{2}{*}{$\lambda_{sep}$} & \multirow{2}{*}{$\lambda_{dist}$} & \multicolumn{3}{c}{mIoU (\%)} \\
        & & & Old & New & All \\
        \midrule
        0.1 & 0.2 & 0.1 & 74.5 & 65.8 & 71.0 \\
        0.3 & 0.2 & 0.1 & 74.8 & 65.5 & 71.2 \\
        \midrule
        0.2 & 0.1 & 0.1 & 73.2 & 64.0 & 69.8 \\
        0.2 & 0.4 & 0.1 & 74.9 & 65.2 & 71.1 \\
        \midrule
        0.2 & 0.2 & 0.05 & 73.5 & \textbf{66.9} & 70.9 \\
        0.2 & 0.2 & 0.25 & \textbf{75.5} & 64.8 & 71.3 \\
        \midrule
        \rowcolor{gray!20}
        \textbf{0.2} & \textbf{0.2} & \textbf{0.1} & 75.1 & 66.3 & \textbf{71.7} \\
        \bottomrule
    \end{tabular}
    \label{tab:sensitivity}
\end{table}

\textbf{Sensitivity Analysis of Confidence Threshold $\tau$.}
Table \ref{tab:ablation_tau} examines the impact of the SLA confidence threshold $\tau$, showing optimal performance (71.7\% mIoU) at $\tau=0.6$. 
Lower thresholds (e.g., 0.4) introduce excessive noise from CAMs, while higher values (e.g., 0.8) result in sparse supervision that hinders new feature learning. 
Thus, $\tau=0.6$ effectively balances noise suppression with supervision sufficiency.

\begin{table}[t]
    \centering
    \caption{Impact of confidence threshold $\tau$ in SLA. 
    We report mIoU for old (1-10), new (11-20), and all classes on the VOC 10-5 setting. 
    The default configuration is \colorbox{gray!20}{highlighted in gray}.}
    \label{tab:ablation_tau}
    \setlength{\tabcolsep}{3.5mm} 
    \begin{tabular}{c|ccc}
        \toprule
        $\tau$ & mIoU (Old) & mIoU (New) & mIoU (All) \\
        \midrule
        0.4 & 74.3 & 63.7 & 70.0 \\
        0.5 & 75.0 & 64.9 & 70.9 \\
        \rowcolor{gray!20}
        \textbf{0.6} & \textbf{75.1} & \textbf{66.3} & \textbf{71.7} \\
        0.7 & 74.0 & 65.2 & 70.6 \\
        0.8 & 73.3 & 64.3 & 69.8 \\
        \bottomrule
    \end{tabular}
\end{table}

\begin{figure}[] 
    \centering
    \includegraphics[width=\linewidth]{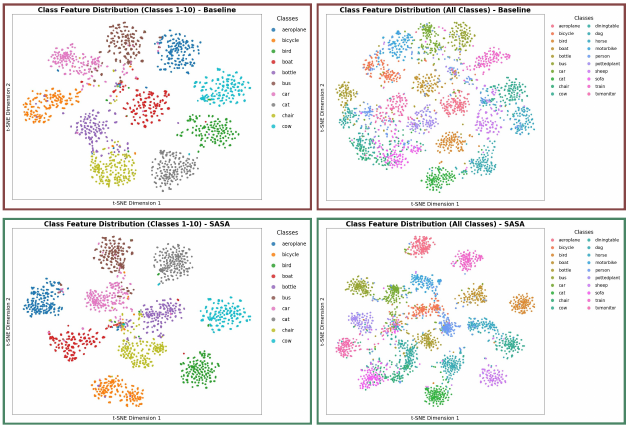}
    \caption{t-SNE comparison between \textit{Baseline} (top) and SASA (bottom) on VOC 10--5. Left: feature clusters of old classes (1--10) after the initial step. Right: clusters of all classes (1--20) upon completion of the incremental task.}\label{fig:tsne}
\end{figure}

\textbf{t-SNE Visualization of Feature Space.}
To evaluate the quality of learned representations, we visualize pixel-level feature distributions using t-SNE in Fig.~\ref{fig:tsne}. 
The Baseline (top row) suffers from severe catastrophic forgetting, where old-class features (1--10) become scattered and overlap with new categories (11--20). 
In contrast, SASA preserves more compact old-class clusters and clearer boundaries between categories. 
This shows that SASA mitigates feature drift and learns more discriminative representations for both old and new classes.

\begin{figure}[] 
    \centering
    \includegraphics[width=\linewidth]{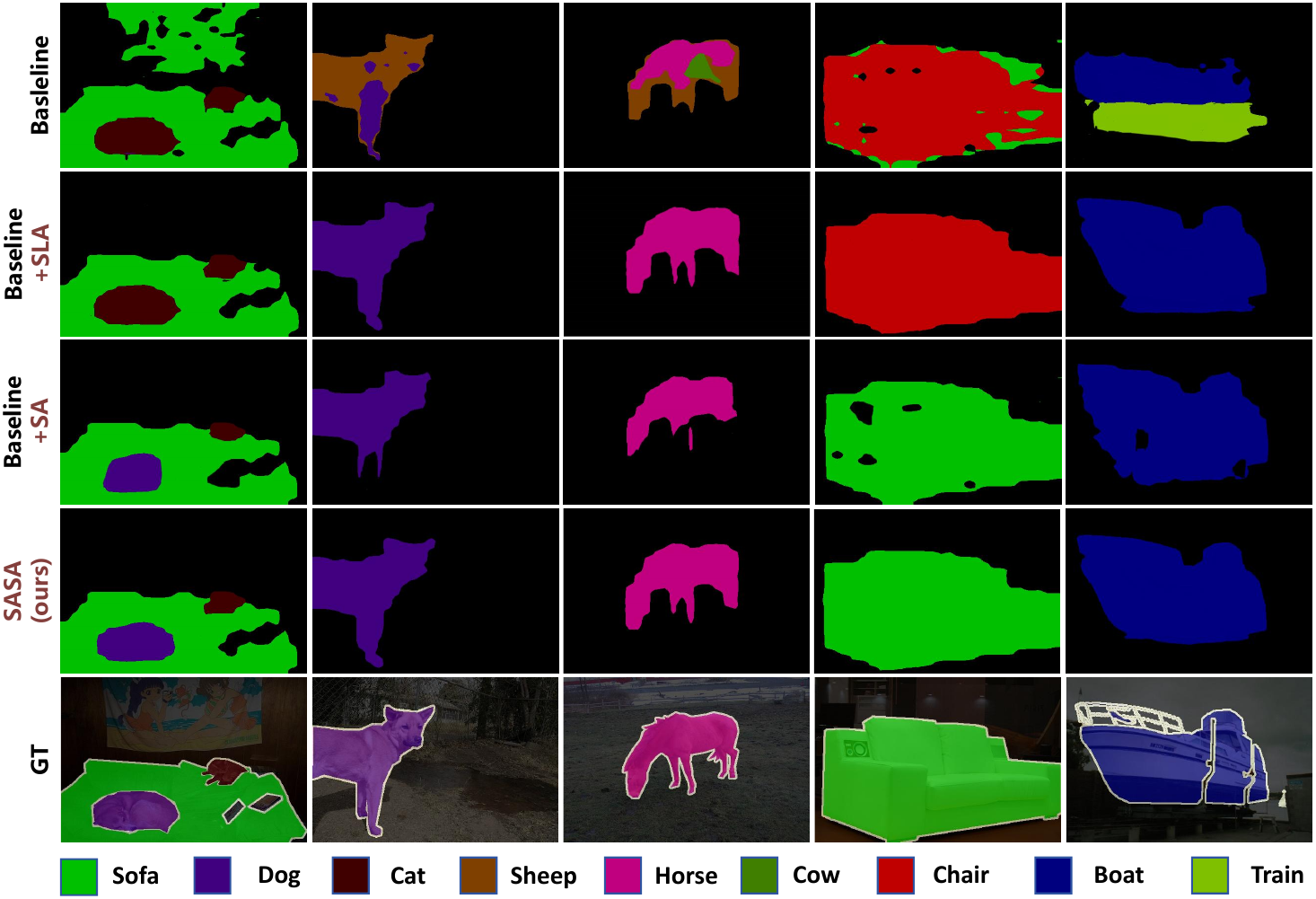}
    \caption{Qualitative segmentation results on the VOC 10-5 task. 
    SASA significantly mitigates feature drift and yields more accurate boundaries compared to the \textit{Baseline}.}
    \label{fig:qual}
\end{figure}

\textbf{Qualitative Segmentation Results.}
Visual comparisons (refer to Fig.~\ref{fig:qual}) demonstrate that SASA significantly mitigates class confusion (\textit{e.g.}, horse vs. sheep) and resolves semantic fragmentation (\textit{e.g.}, sofa misclassified as chair). 
By enforcing the ``One Object, One Class'' constraint, SASA produces sharper boundaries and more coherent segmentation masks compared to the baseline.



%% file: paper_section/conclusion.tex
This paper addresses Feature Drift in WILSS caused by accumulating supervision noise. 
We propose SASA, a drift-resilient framework that stabilizes learning through two key designs: 
(i) Rigid Semantic Anchors coupled with Elastic Residual Adaptation to preserve class identity while accommodating instance variations; 
and (ii) Spatial Label Arbitration (SLA) to filter noisy signals via geometry-aware consistency. 
Extensive evaluations on Pascal VOC and MS COCO show that SASA sets new state-of-the-art performance, especially in multi-step scenarios, effectively mitigating feature drift in weakly supervised settings.